\documentclass[journal]{IEEEtran}
\usepackage{cite}

\ifCLASSINFOpdf
\else
\fi
\usepackage{algorithm}
\usepackage{algorithmic}
\usepackage{amsmath}
\usepackage{amsfonts}
\usepackage{graphicx}
\usepackage{float}
\usepackage{booktabs}
\usepackage{multirow}
\usepackage[table,xcdraw]{xcolor}
\usepackage{overpic}
\usepackage{lineno}  
\begin{document}
%
\title{DMRVisNet: Deep Multi-head Regression Network for Pixel-wise Visibility Estimation \\ Under Foggy Weather}
%
%
%

\author{Jing YOU*,~\IEEEmembership{}
        Shaocheng JIA*,~\IEEEmembership{}
        Xin PEI,~\IEEEmembership{}
        and Danya YAO~\IEEEmembership{}
\thanks{* Contributed equally.}
\thanks{This work has been submitted to the IEEE for possible publication. Copyright may be transferred without notice, after which this version may no longer be accessible. The work described in this paper was supported by the National Key Research and Development Program of China (Grant No. 2021YFC3001501) and the National Natural Science Foundation of China (Grant No. 71671100). (Corresponding author: Xin PEI)}
\thanks{Jing YOU is with the Department of Automation, Tsinghua University, Beijing, China; Shaocheng JIA is with the Department of Civil Engineering, The University of Hong Kong, Hong Kong, China; Xin PEI is with the Department of Automation, Beijing National Research Center for Information Science and Technology, Tsinghua University, Beijing, China; Danya YAO is with the Department of Automation, Beijing National Research Center for Information Science and Technology, Tsinghua University, Beijing, China.}
}

\maketitle

\begin{abstract}
Scene perception is essential for driving decision-making and traffic safety. However, fog, as a kind of common weather, frequently appears in the real world, especially in the mountain areas, making it difficult to accurately observe the surrounding environments. Therefore, precisely estimating the visibility under foggy weather can significantly benefit traffic management and safety. To address this, most current methods use professional instruments outfitted at fixed locations on the roads to perform the visibility measurement; these methods are expensive and less flexible. In this paper, we propose an innovative end-to-end convolutional neural network framework to estimate the visibility leveraging Koschmieder’s law exclusively using the image data. The proposed method estimates the visibility by integrating the physical model into the proposed framework, instead of directly predicting the visibility value via the convolutional neural work. Moreover, we estimate the visibility as a pixel-wise visibility map against those of previous visibility measurement methods which solely predict a single value for an entire image. Thus, the estimated result of our method is more informative, particularly in uneven fog scenarios, which can benefit to developing a more precise early warning system for foggy weather, thereby better protecting the intelligent transportation infrastructure systems and promoting its development. To validate the proposed framework, a virtual dataset, FACI, containing 3,000 foggy images in different concentrations, is collected using the AirSim platform. Detailed experiments show that the proposed method achieves performance competitive to those of state-of-the-art methods.
\end{abstract}. 
    

%
\IEEEpeerreviewmaketitle

\section{Introduction}
%
%
%
%

\IEEEPARstart{V}{isibility}, as an important indicator in traffic management, is defined as the maximum horizontal distance at which a person with normal vision can recognize a black and moderately sized target object from the sky background by the World Meteorological Organization, International Commission on Illumination (CIE), and American Meteorological Society \cite{palvanov2019visnet, li2019method, rasouli2016measurement}. According to the level of degradation, visibility is classified into 10 classes by the British Meteorological Service, in which the lowest level corresponds to heavy fog with visibility less than 50 meters while the highest level corresponds to the visibility greater than 30 kilometers \cite{palvanov2019visnet}. 

Fog, as one of the most common contributing factors to low visibility, is defined as a kind of cloud on the ground and formed by the suspension of microscopic moisture dewdrops into airborne particles \cite{chaabani2017neural}. The low visibility caused by foggy weather will lead to various consequences. For instance, low visibility will pose a threat to traffic safety by making it difficult for drivers to perceive the driving environment and further make driving decisions \cite{chaabani2017neural, kwon2004atmospheric, babari2011model, babari2011computer}. Previous studies have validated that the traffic accident rate strikingly increases when the visibility is less than 200 meters \cite{cheng2018variational}, even the fatal multi-vehicle pileups \cite{kwon2004atmospheric, pomerleau1997visibility}. More specifically, approximately 30\% of fatal crashes were related to fog in the USA from 1990 to 2012 and almost two-thirds of them led to serial collisions involving 10 or more vehicles \cite{hamilton2014hidden}. These crashes can heavily damage the transportation infrastructure systems. Moreover, visibility is also correlated with flight safety \cite{fabbian2007application, costa2006fog, zhu2017application}, image quality \cite{sakaridis2018semantic, slawinski1991neural}, air quality \cite{hyslop2009impaired, thach2010daily}, even economy \cite{pejovic2009factors, dietz2019forecasting, li2017meteorological}.

Apart from the transportation infrastructure systems probably being damaged by the low visibility-induced accidents, some equipment and sensors might also be affected by the low visibility weather, such as camera and Light Detection And Ranging (LiDAR)). Thus, obtaining accurate visibility information can protect the transportation infrastructure systems, as well as promote the development of intelligent transportation infrastructure systems.

To this end, it is necessary to precisely estimate the visibility under foggy weather. However, it is extremely challenging because (1) visibility is a rather complex parameter and influenced by many atmospheric conditions, such as light scatter, air light, and light absorption; (2) using a single value to represent visibility is difficult unless the fog is uniform across the whole image, which occurs rarely in reality \cite{palvanov2019visnet, kwon2004atmospheric}. Still, the earliest literature concerning visibility estimation can date back to the 1920s when Koschmieder’s law \cite{koschmieder1924theorie}, a fundamental theory for atmospheric visibility estimation, was proposed. Following that, tons of work has been done using different methods, including the traditional methods, the statistical methods, and the recent deep neural network (DNN) methods.

Traditional methods adopt either manual observations or professional instruments. For observations, the result may change from observer to observer; For professional instruments, they are costly and less flexible \cite{chaabani2017neural, pomerleau1997visibility}. By contrast, the statistical methods estimate the visibility from the collected data using either the definition or the relationship between the visibility and the data. These methods also require geographic calibration or solely can be used in some special cases. Recently, Some researchers applied DNN models to performing visibility estimation. However, DNN models require large-scale datasets for training, while collecting data from real-world situations is time-consuming and expensive. Importantly, solely using DNN models to estimate the visibility in an end-to-end manner is difficult to interpret how the models work \cite{palvanov2019visnet, palvanov2018dhcnn, you2018relative}. Additionally, publicly available visibility estimation datasets are very rare because collecting visibility data in the real world is time-consuming and costly, which hinders the development of visibility estimation methods.

To tackle the aforementioned issues, we propose a novel multi-head regression convolutional neural networks (CNN) framework, DMRVisNet, which integrates the physical laws and the deep learning methods to estimate the physical parameters in Koschmieder’s law \cite{koschmieder1924theorie} instead of directly predicting visibility. Moreover, the proposed method outputs a visibility map, for which it is more robust and informative under uneven fog situations. To validate the proposed method, a virtual dataset, Foggy Airsim City Images (FACI) is collected using the Airsim simulation platform \cite{shah2018airsim}.

To sum up, our contributions are as follows:
\begin{itemize}

\item We propose a novel multi-head regression convolutional neural network framework, DMRVisNet, to estimate visibility by integrating the physical laws and the deep learning methods;

\item We propose a new pixel-wise visibility estimation paradigm, which is more informative and practical, particularly in uneven fog scenarios;

\item We propose a new virtual dataset, FACI, to be used for both pixel-wise and single-value visibility estimation (it will be publicly available once the paper is accepted).

\item The proposed method achieves performance competitive compared to those of state-of-the-art methods.

\end{itemize}

The remainder of this paper is organized as follows. Section II introduces related works. Section III mathematically defines the problem and presents notational conventions. Section IV presents the proposed dataset, model, and loss functions. Section V reports on detailed experiments. Section VI discusses the limitations of the proposed model. Section VII draws the conclusions.

 
\section{RELATED WORK} 
In this section, we review the literature related to visibility estimation, which is divided into 3 categories: the traditional methods, the statistical methods, and the methods based on DNNs.

\subsection{Traditional methods}

Traditional methods mainly consist of two types of estimation paradigms, i.e., manual observation and professional instruments. For the former, the estimation accuracy heavily depends on the observer. For the latter, professional instruments (e.g., scatterometers, transmissometers, and Light Detection And Ranging (LiDAR)) are first used for measuring the environmental parameters, such as extinction coefficient and the transmission, then the Koschmieder’s law \cite{koschmieder1924theorie} is used for visibility estimation. These professional instruments are comparatively precise, yet they are costly and require specialized installation and calibration. Accordingly, professional instruments are solely used in some special cases \cite{chaabani2017neural, pomerleau1997visibility}, such as fixed-location meteorological stations.

\subsection{Statistical methods}


Statistical methods also can be roughly divided into two estimation paradigms, i.e., estimation by definition and correlation between visibility and feature respectively, and both of them are data-driven methods. For definition estimation, it is to estimate the farthest visible object in the scene. For example, Pomerleau et al. detected lane lines in the image captured by the on-board camera and then estimated visibility leveraging the attenuation of contrast \cite{pomerleau1997visibility}. For correlation estimation, it is to discover the underlying relationship between visibility and features extracted from source data (e.g., images). For instance, Hallowell et al. attempted to find the relationship between visibility and edge attenuation using the images captured by the fixed-position camera in sunny weather \cite{hallowell2007automated}. Following that, so many works have been done in improving either feature extraction or correlation function \cite{xie2008estimating, liaw2010using, babari2012visibility, babari2011model, he2010single, wauben2016exploration, dietz2019forecasting, cheng2018expressway, li2017visibility}.

Statistical methods are much more convenient than traditional methods. However, most statistical methods require either geographic calibration or special scenarios to be used, which hinders statistical methods being widely used.

\subsection{Methods based on DNNs}


Recent performance advances in a variety of computer vision tasks are own to the development of deep neural networks to great extent. Thus, there are also so many trials applying the DNNs to Visibility estimation. In 1995, multilayer perceptron (MLP) was first introduced to perform visibility estimation \cite{pasini1995short}. Following that, several MLP-based methods were introduced \cite{wang2009risk, chaabani2017neural}. Representatively, the risk neural network was proposed to accommodate the basic principle that low visibility should have a higher risk value for the same forecast error \cite{wang2009risk}, which improved the performance of the standard MLP method.

Thereafter, convolutional neural networks (CNNs) were prevalent because of their excellent performance in image processing. Therefore, tremendous work has been conducted by introducing CNNs in visibility estimation \cite{li2017meteorological, specht1991general,you2018relative,palvanov2018dhcnn,palvanov2019visnet}. In these methods, CNNs are used to either extract features from the input images or learn the correlations between the features and visibility. Additionally, the abundant features of multimodal data (e.g., ordinary camera and infrared camera) were used to improve the performance of visibility estimation as well \cite{wang2020multimodal}. However, few of them consider integrating the physical laws for better interpretation and performance improvement.

In summary, traditional methods have solid mathematical foundations but are very difficult to be widely used taking the capital cost and flexibility into account. Statistical methods are early explorations of the data-driven paradigm. Current deep neural network methods are capable of extracting the deep and abstract features from the raw data, but lack interpretability. To the best of our knowledge, none of the work attempts to integrate the physical laws and deep learning methods; none of the work proposes pixel-wise visibility estimation paradigm, either, however, which is significant for uneven fog cases. These issues motivate us to conduct this work.


\section{PROBLEM SETUP}
\label{setup}

Our pixel-wise visibility estimation system $\Psi$ comprises three parts, namely the airlight estimation network, the transmission estimation network, and the depth estimation network, denoted as $\Psi_A$, $\Psi_T$, and $\Psi_D$, respectively. Given an image $I$, the airlight estimation network $\Psi_A$ solely takes the input image $I$ as the input to predict its airlight; this can be mathematically defined as: $\Psi_A: I \in {\mathbb{R}^{h\times w \times c}} \to A \in \mathbb{R}^{3}$, where $A$, $h$, and $w$ are the predicted airlight of the input image $I$, the height of the input image, and the width of the input image, respectively. Similarly, the transmission estimation network $\Psi_T$  takes the input image $I$ as the input to predict its transmission map; this can be mathematically defined as: $\Psi_T: I \in {\mathbb{R}^{h\times w \times c}} \to T \in \mathbb{R}^{h \times w}$, where $T$ is the predicted transmission map of the input image $I$. To make the network easier to be optimized, $\Psi_D$, which is used to estimate the depth map, will take the input image $I$ as input and output the disparity map $\bar{D}$, i.e. $1 \oslash D$, where $D$ and $\oslash$ represent the predicted depth map of the input image $I$ and element-wise division, respectively; this can be mathematically defined as: $\Psi_D: I \in {\mathbb{R}^{h\times w \times c}} \to \bar{D} \in \mathbb{R}^{h \times w}$.

Theoretically, given the transmission map and depth map of the input image, the pixel-wise visibility map can be calculated by Koschmieder’s law \cite{koschmieder1924theorie}, which is mathematically defined as Eq. \ref{problem}.
\begin{equation}
\label{problem}
  V = \ln(\epsilon) \otimes D \oslash \ln(T),
\end{equation}
where $V \in \mathbb{R}^{h \times w}$ and $\otimes$ represent the visibility map and element-wise multiplication respectively; $\epsilon$ is the threshold  defined  by  the  International  Commissionon Illumination (CIE) to describe the meteorological visibility distance. Note that the airlight $A$ is used to train the model as an aid, though it is not shown in the Eq. \ref{problem}.


\section{METHOD}
In this section, visibility estimation theories used for generating the FACI dataset and designing DMRVisNet are first introduced. Subsequently, FACI dataset and DMRVisNet are introduced, respectively. Finally, the loss functions used in this paper are presented.


\subsection{Visibility estimation}

The intensity of the light will gradually decrease along with the increase of the distance due to the scattering, refraction, and other physical phenomenons. In reality, we can find that the luminance of the further object is weaker than that of the closer one, even they have exactly identical luminance in fact. This phenomenon can be mathematically described as Eq. \ref{k1} \cite{koschmieder1924theorie}.
\begin{equation}
\begin{aligned}
    \hat{I} &= \hat{J}\hat{T} + \hat{A}(1-\hat{T}) \\
    \hat{T} &= e^{-\hat{\beta}\hat{D}},
    \label{k1}
\end{aligned}
\end{equation}
wherein $\hat{J}$, $\hat{I}$, $\hat{T}$, $\hat{A}$, $\hat{D}$, and $\hat{\beta}$ represent the intrinsic luminance of the object, the observed luminance of the object, the transmission at the position of the object, the airlight (it results from daylight scattered by the slab of fog between the object and the observer), the depth of the object, and the extinction coefficient of the atmosphere, respectively. 

According to Eq. \ref{k1}, we can learn that the observed luminance is the linear weighting between the intrinsic luminance and the airlight; the weighting factor, $\hat{T}$, decays exponentially against the product of the depth and the extinction coefficient.

Similarly, for a digital image, applying Eq. 2 we can readily deduce to Eq. \ref{k1_mat}.
\begin{equation}
\begin{aligned}
    I &= J \otimes T + A \otimes (1-T)\\
    T &= e^{\beta \otimes D},
    \label{k1_mat}
\end{aligned}
\end{equation}
where $J$, $I$, $T$, $A$, $D$, and $\beta$ represent the fogless image, the foggy image, the transmission map, the airlight map, the depth map, and the extinction coefficient map, respectively. Generally, the airlight across the different positions in the image is regarded as constant \cite{nayar1999vision, narasimhan2003contrast, fattal2008single, cai2016dehazenet, ju2017visibility}. Thus, the airlight map is single-valued.

To define the attenuation of the contrast, the definition of contrast is first introduced, being the relative luminance of the object compared to its background, which is stated in Eq. \ref{contrast}.
\begin{equation}
\label{contrast}
Con = \frac{Obj - Back}{Back},
\end{equation}
where $Con$, $Obj$, and $Back$ represent the contrast of the object, the luminance of the object, and the luminance of the background, respectively.

Therefore, consolidating Eq. \ref{k1} and Eq. \ref{contrast} we can readily derive to the attenuation of the contrast \cite{middleton1957vision}, which can be described as \ref{k2}.
\begin{equation}
\begin{aligned}
    C_{\hat{J}} &= \frac{\hat{J}-\hat{A}}{\hat{A}}\\
    C_{\hat{I}} &= C_{\hat{J}} \hat{T},
    \label{k2}
\end{aligned}
\end{equation}
similarly, $C_{\hat{J}}$ represents contrast between the intrinsic luminance of object and the airlight; $C_{\hat{I}}$ represents contrast between the observed luminance of the object and the airlight. Eq. \ref{k2} indicates that the contrast between the luminance of the object and the airlight decays exponentially against the product of the depth and the extinction coefficient as well.

\begin{table*}[ht]
    \centering
    \caption{Description of data in FACI before augmentation}
    \label{data}
    \begin{tabular}{cc}  
    \toprule   
    Type & Description \\  
    \midrule   
    Scene & Fog-free image   \\
    DepthPerspective & Pixel-wise depth map, you get depth from camera using a projection ray that hits that pixel    \\
    \bottomrule  
    \end{tabular}
\end{table*}

Assuming that the object is black, i.e., $\hat{J} = 0$, thus, Eq. \ref{k2} can be rewritten as Eq. \ref{k3}.
\begin{equation}
\begin{aligned}
    C_{\hat{J}} &= \frac{0-\hat{A}}{\hat{A}} = -1\\
    C_{\hat{I}} &= C_{\hat{J}} \hat{T} = -T.
    \label{k3}
\end{aligned}
\end{equation}
If $|C_{\hat{I}}| < \epsilon$ holds, this black object is considered as unrecognizable, wherein $|\cdot|$ represents absolute value function and $\epsilon$ is the threshold defined by the International Commission on Illumination (CIE) to describe the meteorological visibility distance, generally being 5\%. Accordingly, if $|C_{\hat{I}}| = \epsilon$, visibility $\hat{V}$ is exactly equal to $\hat{D}$. On this basis, we can 
deduce to Eq. \ref{k4} from Eq. \ref{k1} and Eq. \ref{k3}.
\begin{equation}
|C_{\hat{J}}| = |-T| = |-e^{-\hat{\beta} \hat{D}}| = |-e^{-\hat{\beta} \hat{V}}| = e^{-\hat{\beta} \hat{V}} = \epsilon.
\label{k4}
\end{equation}

Applying logarithmic function Eq. \ref{k4} can be rewritten as Eq. \ref{k5}.
\begin{equation}
-\hat{\beta} \hat{V} = \ln(\epsilon).
\label{k5}
\end{equation}
Similarly, applying logarithmic function to the second row of Eq. \ref{k1} we can deduce to Eq. \ref{k6}.
\begin{equation}
\ln(\hat{T}) = -\hat{\beta} \hat{D}.
\label{k6}
\end{equation}
Consolidating Eq. \ref{k5} and Eq. \ref{k6}, Eq. \ref{k7} is finally obtained for calculating visibility.
\begin{equation}
\hat{V} = \frac{-\ln (\epsilon)}{\hat{\beta}} = \frac{\ln (\epsilon) \hat{D}}{\ln (\hat{T})}.
\label{k7}
\end{equation}
Thus, Eq. \ref{k7_mat} can be used to calculate the visibility map for an image.
\begin{equation}
V = \ln (\epsilon) \otimes D \oslash \ln (T),
\label{k7_mat}
\end{equation}
where $V$, $D$, and $T$ represent the visibility map, the depth map, and the transmission map, respectively. We can learn that visibility estimation can be converted to estimate the depth and transmission maps, which are relatively easier to be predicted.
This will be used to devise the visibility estimation framework.

\subsection{FACI dataset}

In this subsection, we first introduce the simulation platform Airsim \cite{shah2018airsim} and its settings. Moreover, the details regarding the collected data are presented. Finally, the dataset's split is shown.

\subsubsection{Simulation platform}

AirSim \cite{shah2018airsim} is an open-source and cross-platform simulator developed by Microsoft for Artificial Intelligence (AI) research, and supports software-in-the-loop simulation with popular flight controllers such as PX4 \& ArduPilot and hardware-in-loop with PX4 for physically and visually realistic simulations; it can help experiment with deep learning, computer vision, and reinforcement learning algorithms, especially for autonomous vehicles.

During collection, Version 1.3.1 of AirSim is used on Windows; among the several maps provided by developers, a city scene is chosen; to make the cameras move more freely, the Computer Vision mode is used; and the height and width of the images are set to 576 and 1,024, respectively.

\subsubsection{Data details}

\begin{figure}[htbp]
    \centering
    \includegraphics[scale=0.1]{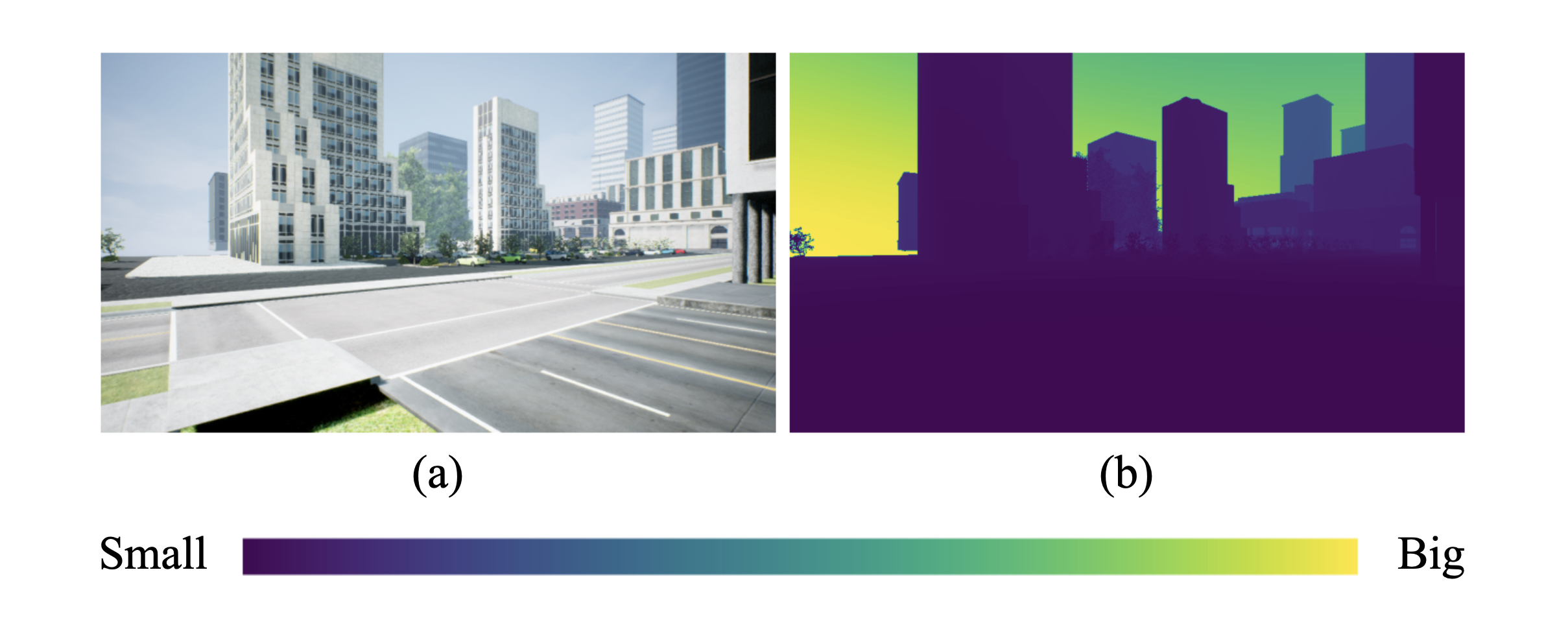}
    \caption{Illustration of the collected original data. (a) Fogless image. (b) Pixel-wise depth map, you get depth from camera using a projection ray that hits that pixel.}
    \label{data1}
\end{figure}

We collected 100 groups of original data in different scenes, and each group consists of fogless images and depth maps. The detailed descriptions regarding the dataset are shown in Table \ref{data} and Figure \ref{data1}. 

\begin{figure}[htbp]
    \centering
    \includegraphics[scale=0.093]{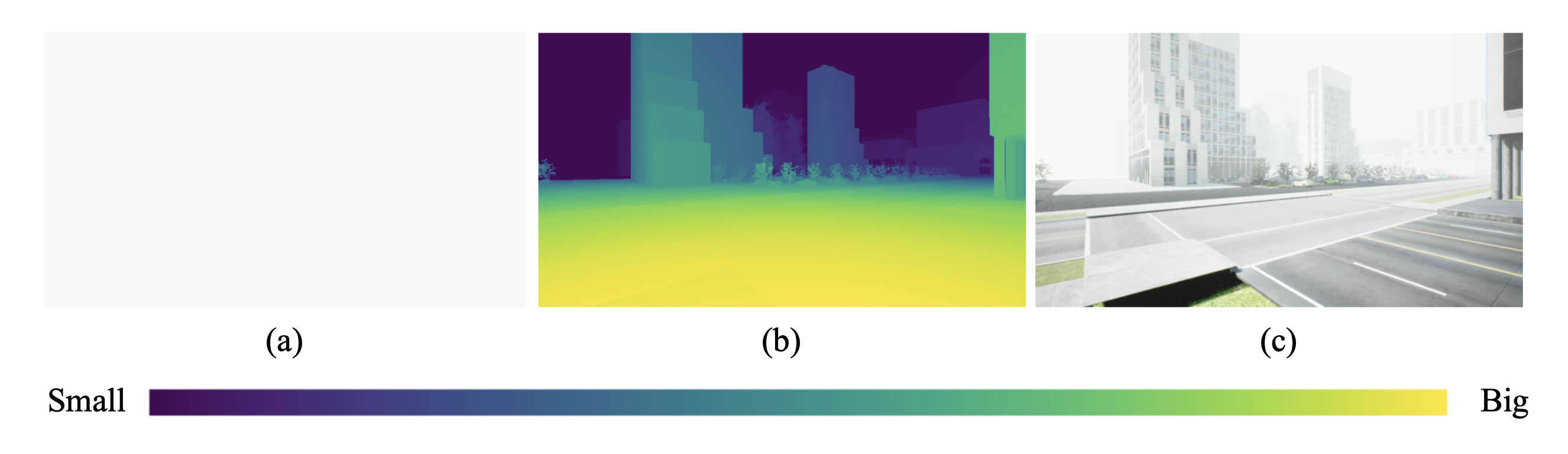}
    \caption{Visualization of augmented data. (a) RGB value of airlight. (b) Pixel-wise transmission map. (c) Foggy image under current airlight and visibility map.}
    \label{data2}
\end{figure}

\begin{table}[htbp]
    \centering
    \caption{Description of additional data in FACI after augmentation}
    \label{data_aug}
    \begin{tabular}{cc}  
    \toprule   
    Type & Description \\  
    \midrule   
    Visibility & Visibility value  \\  
    A & RGB value of airlight  \\    
    T & Pixel-wise transmission map \\
    FoggyScene & Foggy image under current airlight and visibility map\\
    \bottomrule  
    \end{tabular}
\end{table}


Thereafter, we use the collected original data to generate the different foggy images. For each group of original data, 30 groups of augmented data are generated. The detailed procedure for generating a foggy image is described in Algorithm \ref{augment}. 

\begin{figure*}[htbp]
    \centering
    \includegraphics[width=\textwidth]{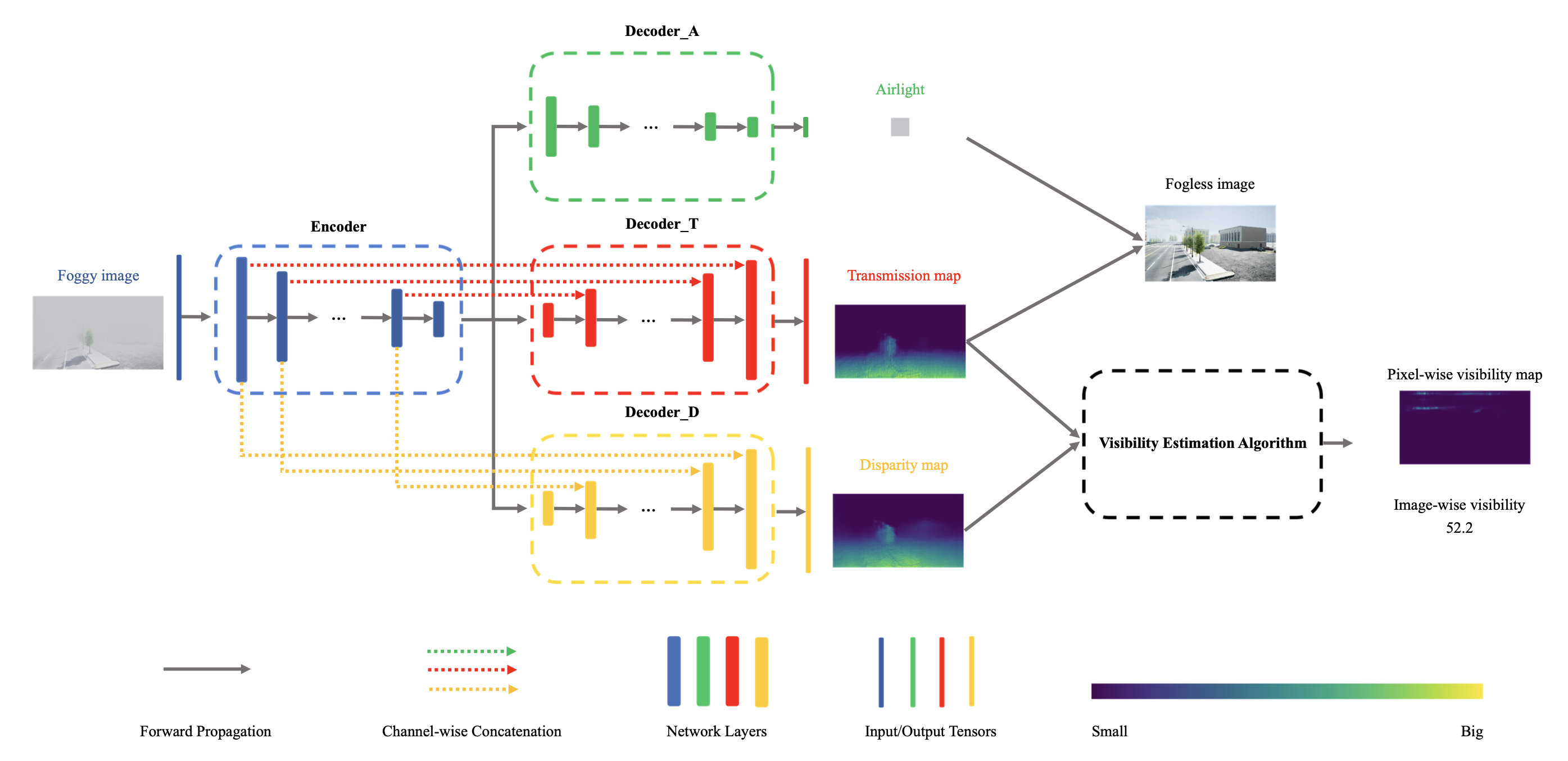}
    \caption{DMRVisNet architecture. ResNet-18 \cite{he2016deep} is adopted as encoder to extract features. Three separate decoders are used to estimate airlight, transmission map, and depth map respectively. Finally, the visibility estimation algorithm is used to couple the outputs to estimate the pixel-wise or image-wise visibility.}
    \label{overview}
\end{figure*}

\begin{algorithm}[htbp]
\caption{Generation of augmented data from original data}  
\label{augment}  
\begin{algorithmic}[1]

\REQUIRE
The fogless image, $J$;
The depth map of fogless image, $D$;

\ENSURE
The visibility map, $V$;
The airlight, $A$;
The transmission map, $T$;
The foggy image under this airlight and visibility map, $I$;  

\STATE To generate uniform fog, sample a value from the uniform distribution $U(10,1000)$ (unit in meter) as the common value of visibility map $V$;

\STATE Generate $T$ with $D$ and $V$ according to Eq. \ref{k7_mat}; 

\STATE Generate the red, green and blue (RBG) values of $A$, while the value of channel B is sampled from $U(180,255)$, the value of channel G is sampled from $\min(U(B-5,B+2), 255)$ and the value of channel R is sampled from $\min (U(\frac{(B+G)}{2}-5,\frac{(B+G)}{2}+2), 255)$, to make the fog appear more real;

\STATE Generate $I$ with $J$, $A$ and $T$ according to Eq. \ref{k1_mat}; 

\RETURN $V$, $A$, $T$, $I$;   

\end{algorithmic}  
\end{algorithm}



The detailed information after augmentation is shown in in Table \ref{data_aug} and Figure \ref{data2}.

\subsubsection{data split}

FACI dataset consists of 3,000 groups data, for which each group contains fogless image, depth map, visibility, airlight, transmission map under current visibility, foggy image under current airlight and visibility. We randomly divide 3,000 images into training set, validation set, and test set with respect to the ratio of 7:2:1, the number of sets of data in training set, validation set and test set respectively is shown in Table \ref{split}. Note that here are no overlapping scenarios between training set, validation set, and test set.

\begin{table}[ht]
    \centering
    \caption{Number of sets of data in training set, validation set and test set}
    \label{split}
    \begin{tabular}{cc}  
    \toprule   
    Type & Number of sets of data \\  
    \midrule   
    Training set & 2100   \\  
    Validation set & 600  \\    
    Test set & 300 \\
    \bottomrule  
    \end{tabular}
\end{table}

\subsection{Model}
\label{model}

\textbf{Model overview.} We propose a novel multi-head regression network, DMRVisNet, to simultaneously predict the airlight, the transmission map, and the depth map. Finally, the visibility estimation algorithm is used to couple the outputs, predicting the pixel-wise or image-wise visibility.

The architecture of DMRVisNet is shown in Figure \ref{overview}, which consists of one encoder and three decoders, named by Encoder, Decoder\_A, Decoder\_T and Decoder\_D, respectively.

\begin{figure*}[htbp]
    \centering
    \includegraphics[width=\textwidth]{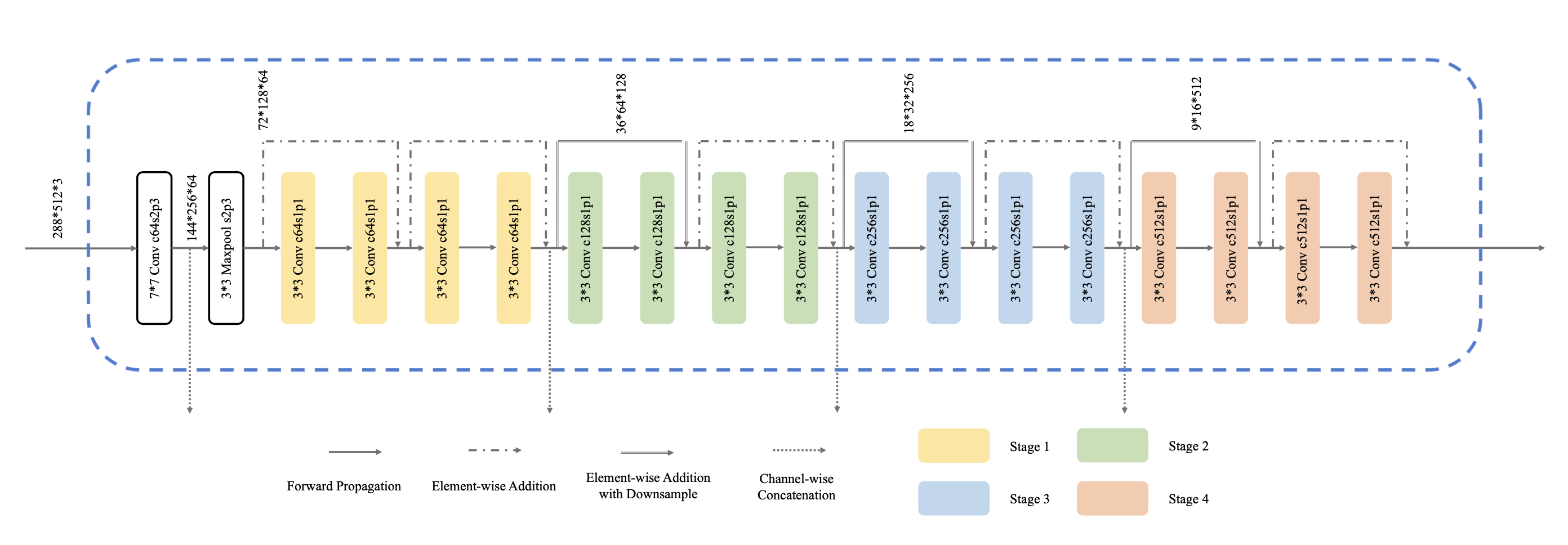}
    \caption{Architecture of Encoder.}
    \label{resnet}
\end{figure*}

\textbf{Encoder.} ResNet-18 \cite{he2016deep} is adopted as encoder to extract multi-scale features. The architecture of the encoder is briefly illustrated in Figure \ref{resnet}, mathematically, which can be described as Eq. \ref{eq_encoder}.
\begin{equation}
F = \text{Encoder}(I),
\label{eq_encoder}
\end{equation}
where $F$ and $I$ represent the features extracted from Encoder and the input image.


\textbf{Decoders.} Following Encoder, three separate decoders, i.e., Decoder\_A, Decoder\_T, and Decoder\_D, are used to estimate airlight, transmission map, and depth map respectively, which can be described as Eq. \ref{eq_decoders}.
\begin{equation}
\begin{aligned}
    \text{Decoder\_A}(F) &= A_{est}\\
    \text{Decoder\_T}(F) &= T_{est}\\
    \text{Decoder\_D}(F) &= \bar{D}_{est},
    \label{eq_decoders}
\end{aligned}
\end{equation}
where $A_{est}$, $T_{est}$, and $\bar{D}_{est}$ represent the estimated airlight, the estimated transmission map and the estimated disparity map respectively. Thus, from an end-to-end manner, the airlight estimation, the transmission map estimation, and the disparity map estimation can be described as Eq \ref{eq_decoders_2}.
\begin{equation}
\begin{aligned}
    \psi_A(I) &= \text{Decoder\_A}(\text{Encoder}(I)) = A_{est}\\
    \psi_T(I) &= \text{Decoder\_T}(\text{Encoder}(I)) = T_{est}\\
    \psi_D(I) &= \text{Decoder\_D}(\text{Encoder}(I)) = \bar{D}_{est},
    \label{eq_decoders_2}
\end{aligned}
\end{equation}
where $\Psi_A$, $\Psi_T$, and $\Psi_D$ represent the airlight estimation network, the transmission estimation network, and the depth estimation network, respectively. 

\begin{figure}[htbp]
    \centering
    \includegraphics[scale=0.14]{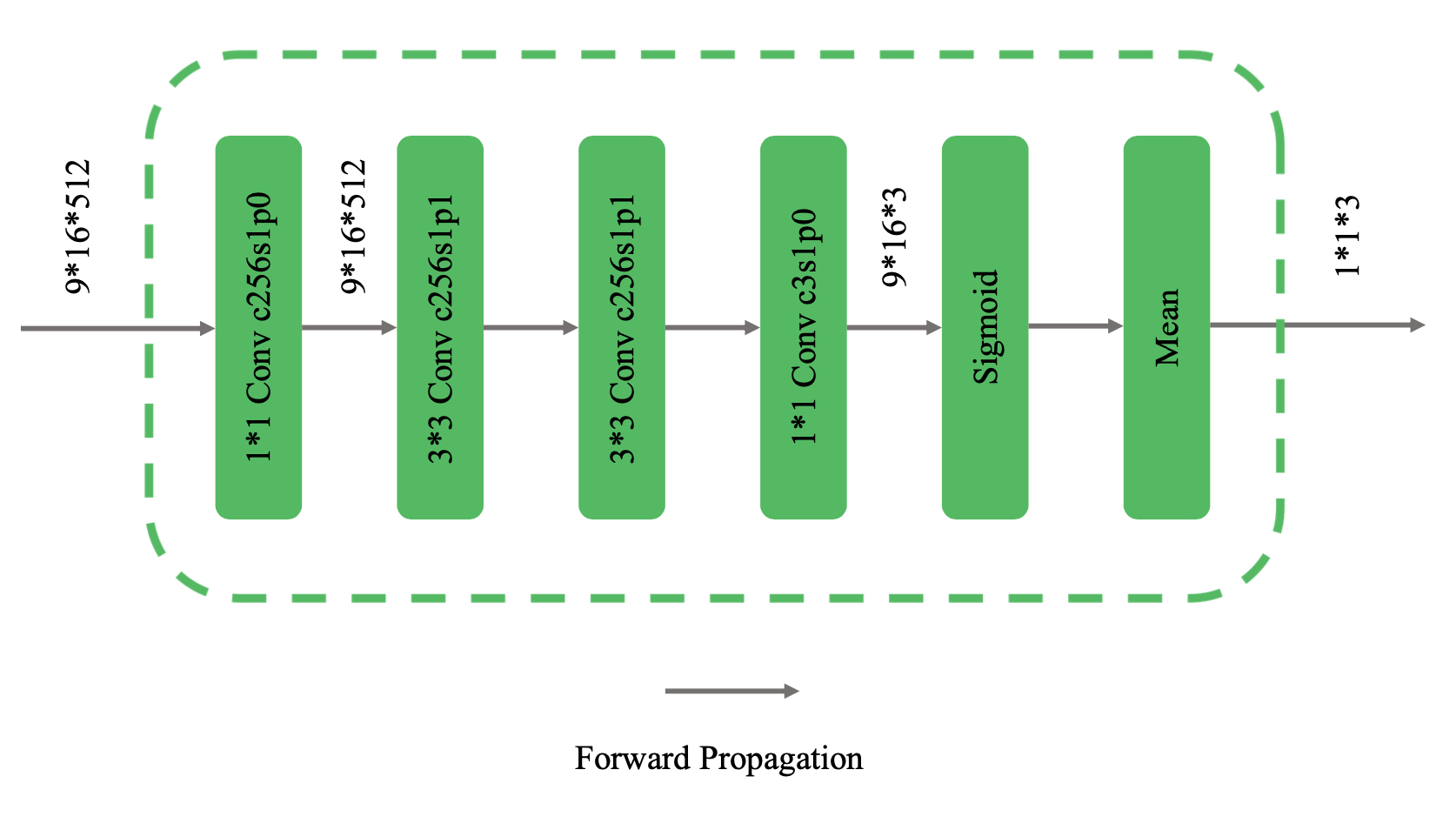}
    \caption{Architecture of Decoder\_A.}
    \label{decoder_a}
\end{figure}

\begin{figure*}[htbp]
    \centering
    \includegraphics[width=\textwidth]{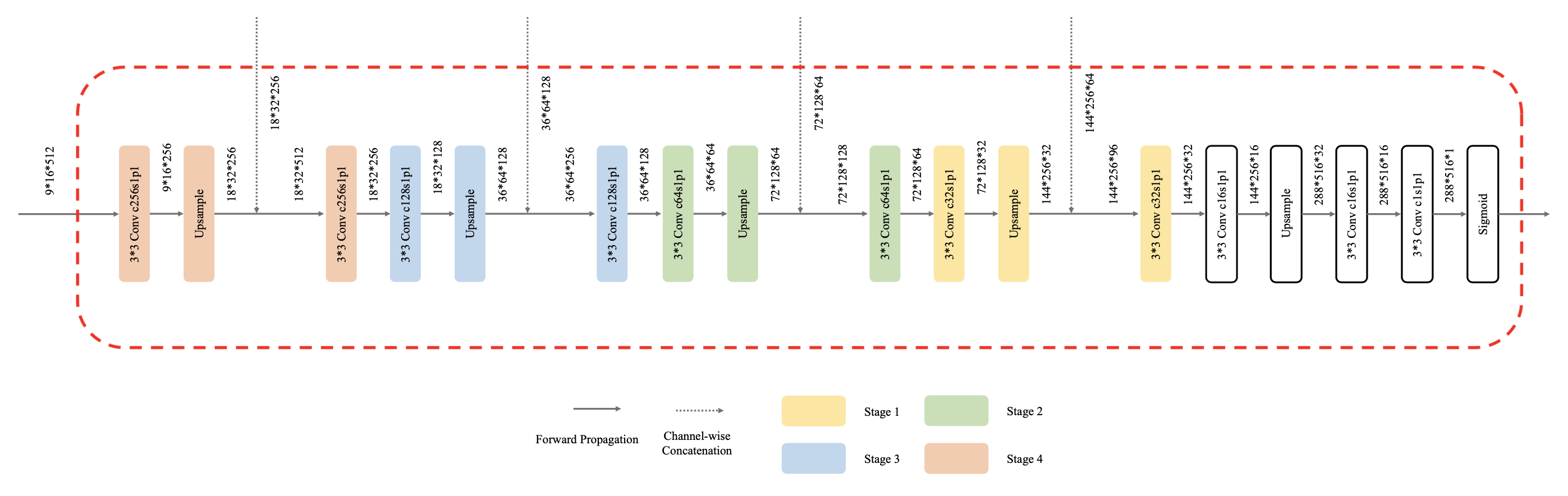}
    \caption{Architecture of Decoder\_T and Decoder\_D.}
    \label{decoder_td}
\end{figure*}

For transmission map and disparity map estimation, a coarse-to-fine architecture used in Monodepth2 \cite{godard2019digging} is adopted, shown as Fig. \ref{decoder_td}. For airlight estimation, several CNN layers are used, shown as Fig. \ref{decoder_a}.


Thus, simply integrating encoder and decoders we can obtain the three results, i.e., airlight, which is used to obtain the estimated fogless image and consequently train the model; transmission map and depth map, which are the inputs of the visibility estimation algorithm.

\begin{algorithm}[htbp]
\caption{Visibility estimation algorithm}  
\label{output}  
\begin{algorithmic}[1]

\REQUIRE
The estimated transmission map, $T_{est}$;
The estimated depth map, $D_{est}$;
The lowest threshold of estimated transmission map $\hat{T}_{min}$;
The highest threshold of pixel-wise visibility map, $\hat{V}_{max}$;
The lowest threshold of pixel-wise visibility map, $\hat{V}_{min}$;

\ENSURE
The estimated pixel-wise visibility map, $V_{est}$;
The estimated image-wise visibility, $\hat{V}_{est}$;

\STATE Generate $V_{est}$ with $T_{est}$ and $D_{est}$ according to Eq. \ref{vis_est};

\STATE Generate $\hat{V}_{est}$ with $V_{est}$ and $T_{est}$ according to Eq. \ref{convert};
\begin{equation}
\begin{aligned}
    &\hat{V}_{est} = 
    \begin{cases}
        \frac{\text{SUM}(V_{est} \otimes mask)}{\text{SUM}(mask)},\quad\text{if}\ \text{SUM}(mask) \neq 0 \\
        \hat{V}_{min},\quad\text{otherwise}
    \end{cases}\\
    &mask = \text{MASK}(\hat{T}_{min}, T_{est}) \otimes \text{MASK}(-\hat{V}_{max}, -V_{est}) \\
    &\text{MASK}(t, mat)(i, j) = 
    \begin{cases}
        1,\quad\text{if}\ mat(i, j) > t \\
        0,\quad\text{otherwise}
    \end{cases}\\
    &\text{SUM}(mat) = \sum_i^h \sum_j^w mat(i,j),
    \label{convert}
\end{aligned}
\end{equation}
where $\text{MASK}$ is a function defined as $\text{MASK}: t \times mat \in \mathbb{R} \times \mathbb{R}^{h\times w} \to \mathbb{R}^{h\times w}$ and $\text{SUM}$ is a function defined as: $\text{SUM}: mat \in \mathbb{R}^{h\times w} \to \mathbb{R}$. Moreover, $i$, $j$, $h$, $w$ represent the row index, the column index, the height and the width of the input matrix $mat$, respectively.

\RETURN $V_{est}$, $\hat{V}_{est}$;   

\end{algorithmic}  
\end{algorithm}

\textbf{Visibility estimation algorithm.} According to Eq. \ref{k7_mat}, we can readily deduce to Eq. \ref{vis_est} for pixel-wise visibility estimation.
\begin{equation}
V_{est} = \ln (\epsilon) \otimes D_{est} \oslash \ln (T_{est}),
\label{vis_est}
\end{equation}
where $V_{est}$, $D_{est}$ and $T_{est}$ represent estimated visibility map, estimated depth map and estimated transmission map, respectively.

To compare with other methods, we also convert pixel-wise visibility to image-wise visibility based on the followings: (1) not every pixel in the visibility map is valid, e.g. pixels in sky area, in which the transmission $\hat{T}$ is 0, the depth $\hat{D}$ is infinity, and the visibility $\hat{V}$ can not be calculated from Eq. \ref{k7}; and 2) some extremely large outliers in visibility map should be ignored due to the limited human vision. Thus, we calculate the mean value of the visibility map where the transmission is greater than the threshold $\hat{T}_{min}$ and the estimated visibility is lower than the maximum visibility $\hat{V}_{max}$. If the number of valid pixels in estimated visibility map is $0$, the visibility is set to the minimum value $\hat{V}_{
min}$, which is $10$ meters for the FACI dataset. The detailed procedure for visibility estimation algorithm is described in Algorithm \ref{output}.


\subsection{Losses}

In this subsection, the loss functions used to train the framework are presented. As presented in Section \ref{model}, the proposed framework first performs multi-task learning, then multiple results are coupled with Koschmieder’s law \cite{koschmieder1924theorie}. Thus, the loss functions for different tasks work together to train the network, which is shown in  Eq. \ref{loss}.
\begin{equation}
\label{loss}
L = \lambda_{A} L_{A}+\lambda_{T} L_{T}+\lambda_{D} L_{D}+\lambda_{\text {defog }} L_{\text {defog }}+\lambda_{\text {vis }} L_{\text {vis }},
\end{equation}
wherein $L$, $L_{A}$, $L_{T}$, $L_{D}$, $L_{defog}$, and $L_{vis}$ represent the final loss function, the loss for airlight estimation, the loss for transmission map estimation, the loss for disparity map estimation, the loss for fogless image estimation, and the loss for visibility estimation, respectively; $\lambda_{A}$, $\lambda_{T}$, $\lambda_{D}$, $\lambda_{defog}$, and $\lambda_{vis}$ are hyperparameters. In what follows, the detailed definition of each 
term in Eq. \ref{loss} is presented. 

\subsubsection{$L_A$}
Root mean square error (RMSE) loss is used for airlight estimation, which can be stated in Eq. \ref{loss_a}.
\begin{equation}
\label{loss_a}
L_A = \| A_{est} - A \|_2,
\end{equation}
where $A_{est}$ represents the estimated airlight, which is derived from the airlight estimation network $\Psi_A$ as Eq. \ref{eq_decoders_2}.

\subsubsection{$L_T$}
Root mean square error (RMSE) loss is also used for transmission map estimation, which can be stated in Eq.  \ref{loss_t}.
\begin{equation}
\label{loss_t}
L_T = \| T_{est} - T \|_2,
\end{equation}
where $T_{est}$ represents the estimated transmission map, which is derived from the transmission estimation network $\Psi_T$ as Eq. \ref{eq_decoders_2}.

\subsubsection{$L_D$}
Following Jia et al. \cite{jia2021self}, we use the combination of L1 Loss, Structure Similarity (SSIM) Loss \cite{wang2004image}, and Smoothness Loss as $L_d$ to train the disparity estimation head, which is stated in Eq. \ref{loss_d}.
\begin{equation}
\begin{aligned}
    L_D &= \lambda_{L 1} \| \bar{D}_{est} - \bar{D} \|_1\\
    &+ \lambda_{SSIM} \frac{1-SSIM(\bar{D}_{est}, \bar{D})}{2}\\
    &+ \lambda_{Smooth} \frac{\text{SUM}(\left|e^{-\partial_{x} (J)}\right| \otimes \partial_{x} (\bar{D}_{est}))}{h*w}\\
    &+ \lambda_{Smooth} \frac{\text{SUM}(\left|e^{-\partial_{y} (J)}\right| \otimes \partial_{y} (\bar{D}_{est}))}{h*w},
    \label{loss_d}
\end{aligned}
\end{equation}
where $\partial$, $\bar{D}$, $\bar{D}_{est}$, $h$, $w$ and $J$ represent the gradient operator, the disparity map, the estimated disparity map, the height of estimated disparity map, the width of estimated disparity map and the fog-free color image, respectively; $\lambda_{L1}$, $\lambda_{SSIM}$, and $\lambda_{Smooth}$ are hyperparameters.

Following Jia et al. \cite{jia2021self}, the hyperparameters mentioned above are set as follows: $\lambda_{L1} = 0.15$, $\lambda_{SSIM} = 0.85$, and $\lambda_{Smooth} = 1e-3$.

\subsubsection{$L_{defog}$}
According to Eq. \ref{k1_mat}, the defogged image $J$ can be obtained from Eq. \ref{defog}.
\begin{equation}
\label{defog}
J = (I-A) \oslash T + A,
\end{equation}
where $I$, $A$, and $T$ represent the foggy image, the airlight and the transmission map, respectively. 

Consequently, the estimated fogless image $J_{est}$ can be obtained from Eq. \ref{defog_est}.
\begin{equation}
\label{defog_est}
J_{est} = (I-A_{est}) \oslash T_{est} + A.
\end{equation}

Then, RMSE loss is used to train the defogging head. This can be mathematically defined as Eq. \ref{loss_defog}.

\begin{equation}
\label{loss_defog}
L_{defog} = \| J_{est} - J \|_2.
\end{equation}

\subsubsection{$L_{vis}$}

Eq. \ref{k7_mat}, which is used to calculate visibility map $V$, can be rewritten as Eq. \ref{k7_mat_2}.
\begin{equation}
V = \ln (\epsilon) \oslash (\bar{D} \otimes \ln (T) ).
\label{k7_mat_2}
\end{equation}
The estimated visibility map $V_{est}$ can be derived from Eq. \ref{k7_mat_3}.
\begin{equation}
\begin{aligned}
V_{est} &= \ln (\epsilon) \oslash (\bar{D}_{est} \otimes \ln (T_{est}) )\\
&= \ln (\epsilon) \oslash (\Psi_D(I) \otimes \ln (\Psi_T(I))).
\end{aligned}
\label{k7_mat_3}
\end{equation}

Note that $\ln (\epsilon)$ is constant, we solely constrain $\Psi_D(I) \otimes \ln (\Psi_T(I))$. To achieve this, L1 Loss is used, which can be defined as Eq. \ref{loss_vis}.
\begin{equation}
\label{loss_vis}
L_{vis} = \| \bar{D}_{est} \otimes \ln (T_{est}) -\bar{D} \otimes \ln (T) \|_1.
\end{equation}

\section{Experiments}

In this section, we first introduce the implementation details. Then the results and comparisons with other state-of-the-art methods are showed. Thereafter, the parameter analysis, ablation studies and model complexities are presented, respectively.

\subsection{Implementation details}

\textbf{Experimental environment.} All experiments were conducted with PyTorch 1.6.0 \cite{paszke2017automatic} on Ubuntu 16.04.6 LTS. The detailed information is shown in Table \ref{environment}.

\begin{table}[ht]
    \centering
    \caption{Experimental Environment}
    \label{environment}
    \begin{tabular}{cc}  
    \toprule   
    Item & Content \\  
    \midrule   
    CPU & Intel Core i7-6900K CPU @ 3.20GHz   \\  
    GPU & NVIDIA GeForce GTX 1080  \\    
    RAM & 128 GB \\
    Operating System & Ubuntu 16.04.6 LTS \\
    Programming Language & Python 3.8.3 \\
    Deep Learning Framework & Pytorch \cite{paszke2017automatic} 1.6.0\\
    CUDA Version & CUDA 10.1 \\
    \bottomrule  
    \end{tabular}
\end{table}

\textbf{Data augmentation.} During training phase, the image and corresponding pixel-wise labels are resized to $288 \times 512$. Following Jia et al. \cite{jia2021self}, we perform data augmentations as follows: randomly cropping with relative size ranging from $0.75$ to $1$, random horizontal flips with a $50$ percent chance, as well as random brightness, contrast, saturation, and hue jitter, with respective ranges of ±$0.2$, ±$0.2$, ±$0.2$, and ±$0.1$.

\textbf{Hyperparameters.} In training phase, we use Adam optimizer \cite{kingma2014adam} with initial learning rate of $1e-5$ and batch size of $16$. We multiply the learning rate by $0.1$ when the fluctuations of $L$ on validation set are less than $1e-4$ for 10 epochs, until the learning rate drops to $1e-8$. We train DMRVisNet for $300$ epochs on a single GeForce GTX 1080 card, and the training process takes about $6$ hours.


\begin{figure*}[htbp]
    \centering
    \includegraphics[width=\textwidth]{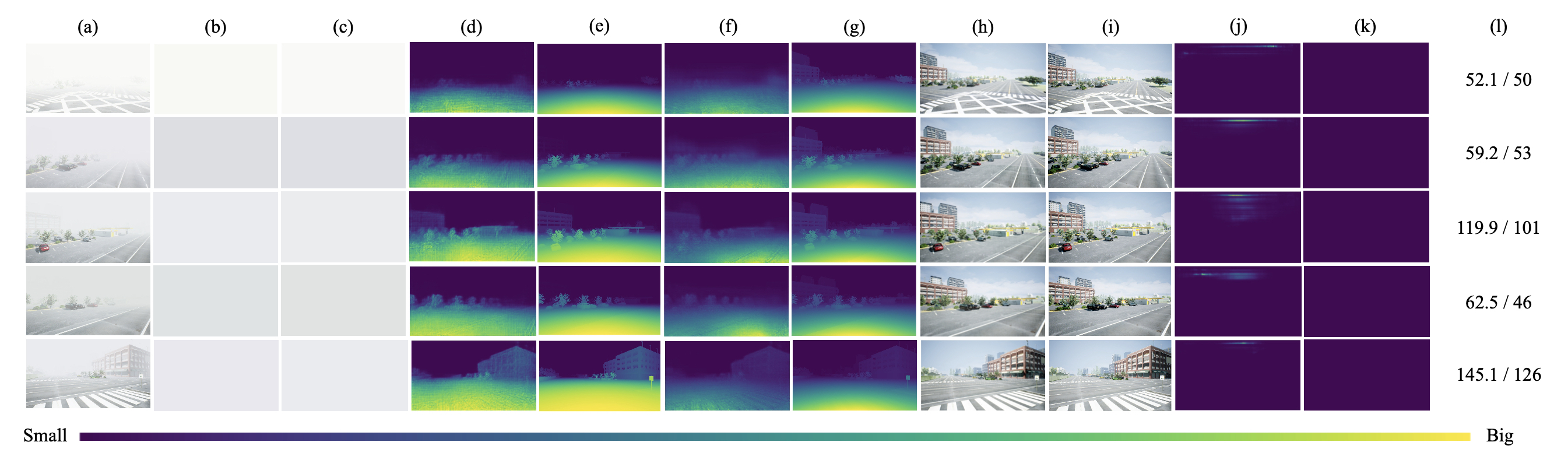}
    \caption{Qualitative results on the test dataset. (a) The input images. (b) The estimated results of airlight. (c) The ground truths of airlight. (d) The estimated results of transmission map. (e) The ground truths of transmission map. (f) The estimated results of disparity map. (g) The ground truths of disparity map. (h) The estimated results of defog image. (i) The ground truths of defog image. (j) The estimated results of pixel-wise visibility map. (k) The ground truths of pixel-wise visibility map. (l) The estimated results of image-wise estimated visibility (left) and the ground truths of image-wise estimated visibility (right).}
    \label{visual_test}
\end{figure*}


\begin{figure}[htbp]
    \centering
    \includegraphics[scale=0.15]{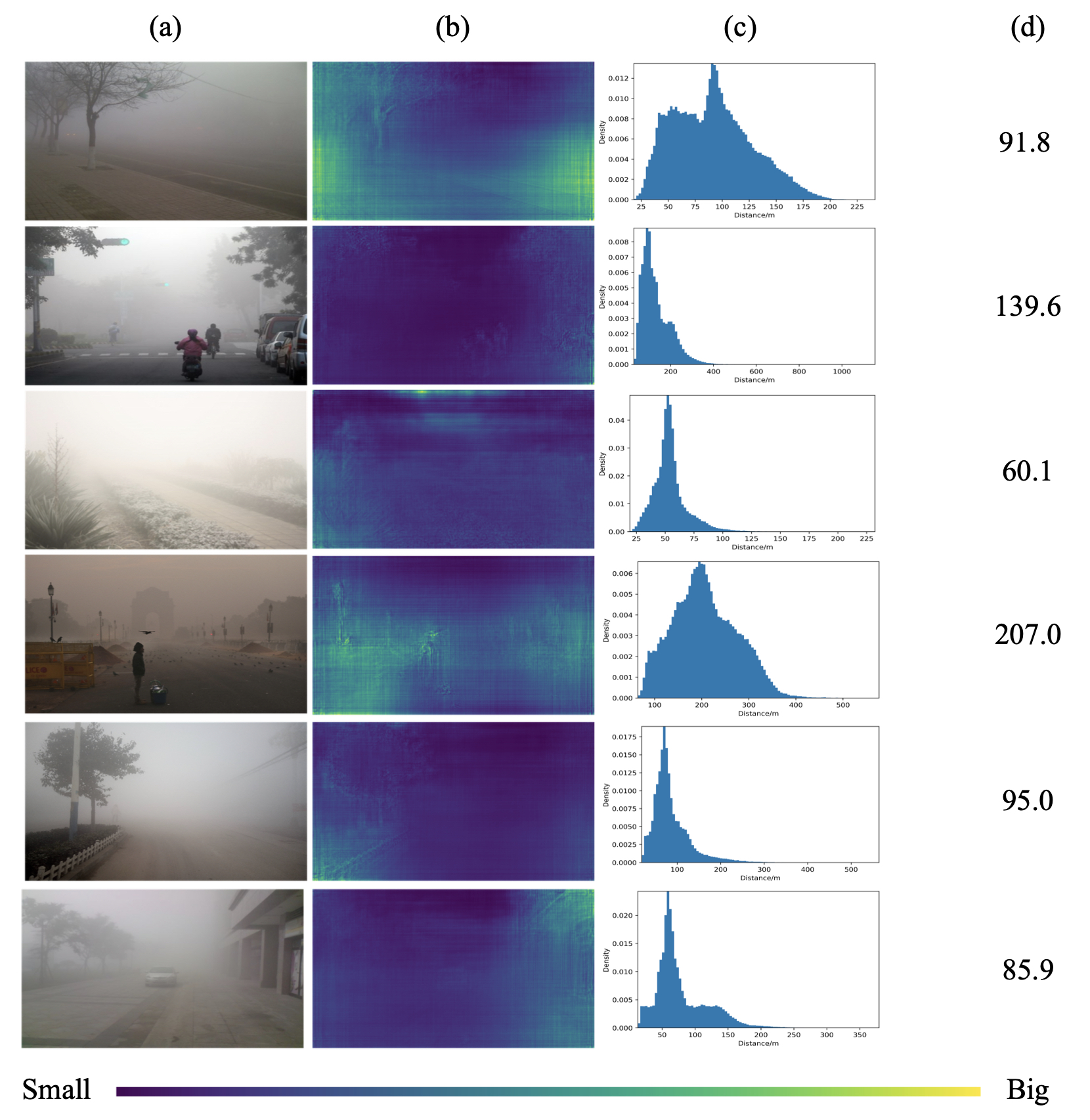}
    \caption{Qualitative results on the real foggy images. (a) The input images. (b) The estimated results of pixel-wise visibility map. (c) The histograms of estimated pixel-wise visibility map. (d) The estimated results of image-wise visibility.}
    \label{visual_real}
\end{figure}

\subsection{Results comparison} 

\subsubsection{Metrics}

For pixel-wise visibility estimation, we use absolute relative error (AbsRel), square relative error (SqRel), root mean square error (RMSE), and root mean square error (RMSElog) as metrics in our experiments. Their definitions are shown in Eq. \ref{absrel}, Eq. \ref{sqrel}, Eq. \ref{rmse}, and Eq. \ref{rmselog}; where $y_{pred}$, $y_{gt}$, and $n$ represent the predicted value, ground truth and the number of valid pixels, respectively.
\begin{equation}
\label{absrel}
\text {AbsRel}=\frac{1}{n} \sum\left|\frac{y_{\text {pred }}-y_{\text {gt }}}{y_{\text {gt }}}\right|.
\end{equation}
\begin{equation}
\label{sqrel}
\text {SqRel}=\frac{1}{n} \sum\left(\frac{y_{p r e d}-y_{g t}}{y_{g t}}\right)^{2}.
\end{equation}
\begin{equation}
\label{rmse}
\text {RMSE}=\sqrt{\frac{1}{n} \sum\left(y_{\text {pred }}-y_{g t}\right)^{2}}.
\end{equation}
\begin{equation}
\label{rmselog}
\text{RMSElog} =\sqrt{\frac{1}{n} \sum\left(\log_{10}\left(y_{\text {pred }}\right)-\log_{10}\left(y_{g t}\right)\right)^{2}}.
\end{equation}
Generally, AbsRel is less than 20\%, the estimated result is acceptable \cite{li2017visibility}.

For image-wise visibility estimation, we use accuracy as the metric. To convert the regression results to classification results, we divide visibility into 5 classes, and each class and its corresponding visibility range are shown in Table \ref{class}. Thus, the continuous outputs can be converted to different classes. The definition of accuracy is shown in Eq. \ref{accuracy_eq}.
\begin{equation}
\label{accuracy_eq}
\begin{aligned}
\text{Accuracy}=\frac{1}{n} \sum \mathcal{I}_{\text{CLASS}(y_{\text{pred}}) \doteq \text{CLASS}(y_{\text{gt}})} \\
\text{CLASS}(y) = 
\begin{cases}
    0,\quad \text{if} \ y < 200\\
    1,\quad \text{if} \ 200 \leq y < 400\\
    2,\quad \text{if} \ 400 \leq y < 600\\
    3,\quad \text{if} \ 600 \leq y < 800\\
    4,\quad \text{if} \ 800 \leq y
\end{cases},
\end{aligned}
\end{equation}
where $\text{CLASS}$ is a function defined as: $\text{CLASS}: y \in \mathbb{R} \to \mathbb{Z}$, $\mathcal{I}$ and $\doteq$ represent indicator function and equivalence relation, respectively.

\begin{table}[htbp]
    \centering
    \caption{Class number and its corresponding visibility range (unit in meter)}
    \label{class}
    \begin{tabular}{cc}  
    \toprule   
    Class number & Visibility Range \\  
    \midrule   
    0 & $<$ 200   \\  
    1 & 200-400  \\    
    2 & 400-600 \\
    3 & 600-800 \\
    4 & $>$ 800 \\
    \bottomrule  
    \end{tabular}
\end{table}

\subsubsection{Results}

\begin{table*}[htbp]
\caption{Quantitative results of the visibility estimation on the FACI dataset. CLS and REG represent classification task and regression task, respectively. The best performances and our model are marked \textbf{bold}. The second best performances are \underline{underlined}.}
\small
\centering
\begin{tabular}{@{}lccccccc@{}}

\toprule

{\color[HTML]{000000} } & 
{\color[HTML]{000000} } & 
{\color[HTML]{000000} } & 
\multicolumn{1}{c}{{\color[HTML]{000000} Errors $\uparrow$}} &
\multicolumn{4}{c}{{\color[HTML]{000000} Errors $\downarrow$}} \\ 

\cmidrule(l){4-8}
\multirow{-2}{*}{{\color[HTML]{000000} Methods}} &
\multirow{-2}{*}{{\color[HTML]{000000} CLS?}} &
\multirow{-2}{*}{{\color[HTML]{000000} REG?}} & 
{\color[HTML]{000000} Accuracy (unit in \%)} &
{\color[HTML]{000000} AbsRel} & 
{\color[HTML]{000000} SqRel} & 
{\color[HTML]{000000} RMSE} & 
{\color[HTML]{000000} RMSElog} \\ 

\midrule

{\color[HTML]{000000} PCASE \cite{chaabani2017neural} } & 
\multicolumn{1}{c}{{\color[HTML]{000000} $\surd$}} &
{\color[HTML]{000000} -} & 
{\color[HTML]{000000} 43.0} &
{\color[HTML]{000000} -} & 
{\color[HTML]{000000} -} & 
{\color[HTML]{000000} -} & 
{\color[HTML]{000000} -} \\

{\color[HTML]{000000} ResNet-50 \cite{he2016deep} } & 
\multicolumn{1}{c}{{\color[HTML]{000000} $\surd$}} &
{\color[HTML]{000000} -} & 
{\color[HTML]{000000} 50.3} & 
{\color[HTML]{000000} -} & 
{\color[HTML]{000000} -} & 
{\color[HTML]{000000} -} & 
{\color[HTML]{000000} -} \\

{\color[HTML]{000000} VGG-16 \cite{simonyan2014very} } & 
\multicolumn{1}{c}{{\color[HTML]{000000} $\surd$}} &
{\color[HTML]{000000} -} & 
{\color[HTML]{000000} \underline{58.3}} & 
{\color[HTML]{000000} -} & 
{\color[HTML]{000000} -} & 
{\color[HTML]{000000} -} & 
{\color[HTML]{000000} -}  \\

{\color[HTML]{000000} VisNet \cite{palvanov2019visnet} } & 
\multicolumn{1}{c}{{\color[HTML]{000000} $\surd$}} &
{\color[HTML]{000000} -} & 
{\color[HTML]{000000} 54.0} & 
{\color[HTML]{000000} -} & 
{\color[HTML]{000000} -} & 
{\color[HTML]{000000} -} & 
{\color[HTML]{000000} -} \\

{\color[HTML]{000000} GRNN \cite{li2017meteorological} } & 
{\color[HTML]{000000} -} & 
\multicolumn{1}{c}{{\color[HTML]{000000} $\surd$}} &
{\color[HTML]{000000} 54.3} & 
{\color[HTML]{000000} \underline{0.27918}} & 
{\color[HTML]{000000} \underline{0.13469}} & 
{\color[HTML]{000000} \underline{156.59050}} & 
{\color[HTML]{000000} \underline{0.14024}} \\

{\color[HTML]{000000} \textbf{Our} } & 
{\color[HTML]{000000} -} & 
\multicolumn{1}{c}{{\color[HTML]{000000} $\surd$}} &
{\color[HTML]{000000} \textbf{68.3}} & 
{\color[HTML]{000000} \textbf{0.17000}} &
{\color[HTML]{000000} \textbf{0.05094}} &
{\color[HTML]{000000} \textbf{93.11345}} &
{\color[HTML]{000000} \textbf{0.08429}} \\

\bottomrule
\end{tabular}
\label{compare}
\end{table*}

For a fair comparison, we reproduce other state-of-the-art methods on our platform using the proposed FACI dataset. All experiments are with identical configurations and hyperparameters. For image-wise visibility classification methods, cross-entropy (CE) loss is used during training. By contrast, Table \ref{class} is used for visibility regression methods. In this case, all methods are comparable.



Table \ref{compare} shows the quantitative results of different methods. We can find that most methods are image-wise classification methods which cannot provide the detailed visibility information. GRNN method \cite{li2017meteorological} is to predict a single continuous visibility value for the input image. However, it cannot sense the uneven fog situations. By contrast, the proposed method can predict a pixel-wise visibility map. To the best of our knowledge, our work is the first to propose the pixel-wise visibility estimation paradigm.

From the numbers presented in Table \ref{compare}, we can learn that the proposed method outperforms other state-of-the-art methods by large margins in all metrics. This demonstrates the effectiveness of the proposed framework.

Fig. \ref{visual_test} shows the qualitative results from the proposed method. We can learn that the proposed method can perform very well in different low-level vision tasks.

To further evaluate the practicality of the proposed method, we apply the model trained on the FACI dataset to some real-world foggy images without any refinement. Fig. \ref{visual_real} presents the qualitative and quantitative results. We can learn that the proposed method can reflect the uniformity of fog from the visibility map or the histograms. For instance, the first and fourth images have uneven fog across different pixel positions, which has been clearly illustrated by the various intensities in visibility map and the bigger variances of the histograms.

To sum up, the quantitative and qualitative results demonstrate the effectiveness and advances of the proposed method. Some qualitative results on real-world data show the strong practicality of the proposed method.

\subsection{Parameter analysis}

\begin{table*}[ht]
\caption{Parameter analysis on $\lambda_A$, $\lambda_T$, $\lambda_D$, $\lambda_{defog}$ and $\lambda_{vis}$. The best performances are marked \textbf{bold}. The second best performances are \underline{underlined}.}
\small
\centering
\begin{tabular}{@{}cccccccccc@{}}

\toprule

{\color[HTML]{000000} } & 
{\color[HTML]{000000} } & 
{\color[HTML]{000000} } & 
{\color[HTML]{000000} } & 
{\color[HTML]{000000} } & 
\multicolumn{1}{c}{{\color[HTML]{000000} Errors $\uparrow$}} &
\multicolumn{4}{c}{{\color[HTML]{000000} Errors $\downarrow$}} \\ 

\cmidrule(l){6-10}
\multirow{-2}{*}{{\color[HTML]{000000} $\lambda_{A}$}} &
\multirow{-2}{*}{{\color[HTML]{000000} $\lambda_{T}$}} &
\multirow{-2}{*}{{\color[HTML]{000000} $\lambda_{D}$}} &
\multirow{-2}{*}{{\color[HTML]{000000} $\lambda_{defog}$}} &
\multirow{-2}{*}{{\color[HTML]{000000} $\lambda_{vis}$}} &
{\color[HTML]{000000} Accuracy (unit in \%)} &
{\color[HTML]{000000} AbsRel} &
{\color[HTML]{000000} SqRel} &
{\color[HTML]{000000} RMSE} &
{\color[HTML]{000000} RMSElog} \\ 

\midrule

{\color[HTML]{000000} $0.6$} & 
{\color[HTML]{000000} $1$} & 
{\color[HTML]{000000} $1$} & 
{\color[HTML]{000000} $1e-6$} & 
{\color[HTML]{000000} $1$} &
{\color[HTML]{000000} \underline{66.0}} & 
{\color[HTML]{000000} 0.19477} & 
{\color[HTML]{000000} 0.06394} & 
{\color[HTML]{000000} \underline{111.68530}} & 
{\color[HTML]{000000} 0.09681} \\

{\color[HTML]{000000} $0.8$} & 
{\color[HTML]{000000} $1$} & 
{\color[HTML]{000000} $1$} & 
{\color[HTML]{000000} $1e-6$} & 
{\color[HTML]{000000} $1$} &
{\color[HTML]{000000} 62.0} & 
{\color[HTML]{000000} \underline{0.18804}} & 
{\color[HTML]{000000} \underline{0.05655}} & 
{\color[HTML]{000000} 114.66917} & 
{\color[HTML]{000000} \underline{0.09305}} \\

{\color[HTML]{000000} $1$} & 
{\color[HTML]{000000} $1$} & 
{\color[HTML]{000000} $1$} & 
{\color[HTML]{000000} $1e-6$} & 
{\color[HTML]{000000} $1$} &
{\color[HTML]{000000} \textbf{67.3}} &
{\color[HTML]{000000} \textbf{0.16662}} & 
{\color[HTML]{000000} \textbf{0.04666}} & 
{\color[HTML]{000000} \textbf{98.05333}} & 
{\color[HTML]{000000} \textbf{0.08452}} \\

{\color[HTML]{000000} $2$} & 
{\color[HTML]{000000} $1$} & 
{\color[HTML]{000000} $1$} & 
{\color[HTML]{000000} $1e-6$} & 
{\color[HTML]{000000} $1$} &
{\color[HTML]{000000} 63.7} & 
{\color[HTML]{000000} 0.23491} & 
{\color[HTML]{000000} 0.10370} & 
{\color[HTML]{000000} 128.72461} & 
{\color[HTML]{000000} 0.11392} \\

{\color[HTML]{000000} $3$} & 
{\color[HTML]{000000} $1$} & 
{\color[HTML]{000000} $1$} & 
{\color[HTML]{000000} $1e-6$} & 
{\color[HTML]{000000} $1$} &
{\color[HTML]{000000} 62.3} & 
{\color[HTML]{000000} 0.24748} & 
{\color[HTML]{000000} 0.10430} & 
{\color[HTML]{000000} 130.80329} & 
{\color[HTML]{000000} 0.11530} \\

\midrule

{\color[HTML]{000000} $1$} & 
{\color[HTML]{000000} $0.6$} & 
{\color[HTML]{000000} $1$} & 
{\color[HTML]{000000} $1e-6$} & 
{\color[HTML]{000000} $1$} &
{\color[HTML]{000000} \underline{63.0}} & 
{\color[HTML]{000000} 0.25336} & 
{\color[HTML]{000000} 0.10989} & 
{\color[HTML]{000000} 132.24173} & 
{\color[HTML]{000000} 0.11595} \\

{\color[HTML]{000000} $1$} & 
{\color[HTML]{000000} $0.8$} & 
{\color[HTML]{000000} $1$} & 
{\color[HTML]{000000} $1e-6$} & 
{\color[HTML]{000000} $1$} &
{\color[HTML]{000000} 61.7} & 
{\color[HTML]{000000} \underline{0.18831}} & 
{\color[HTML]{000000} \underline{0.06276}} & 
{\color[HTML]{000000} \underline{109.26913}} & 
{\color[HTML]{000000} \underline{0.09434}} \\

{\color[HTML]{000000} $1$} & 
{\color[HTML]{000000} $1$} & 
{\color[HTML]{000000} $1$} & 
{\color[HTML]{000000} $1e-6$} & 
{\color[HTML]{000000} $1$} &
{\color[HTML]{000000} \textbf{67.3}} &
{\color[HTML]{000000} \textbf{0.16662}} & 
{\color[HTML]{000000} \textbf{0.04666}} & 
{\color[HTML]{000000} \textbf{98.05333}} & 
{\color[HTML]{000000} \textbf{0.08452}} \\

{\color[HTML]{000000} $1$} & 
{\color[HTML]{000000} $2$} & 
{\color[HTML]{000000} $1$} & 
{\color[HTML]{000000} $1e-6$} & 
{\color[HTML]{000000} $1$} &
{\color[HTML]{000000} 58.7} & 
{\color[HTML]{000000} 0.21396} & 
{\color[HTML]{000000} 0.07158} & 
{\color[HTML]{000000} 110.60130} & 
{\color[HTML]{000000} 0.09903} \\

{\color[HTML]{000000} $1$} & 
{\color[HTML]{000000} $3$} & 
{\color[HTML]{000000} $1$} & 
{\color[HTML]{000000} $1e-6$} & 
{\color[HTML]{000000} $1$} &
{\color[HTML]{000000} 16.7} & 
{\color[HTML]{000000} 0.96148} & 
{\color[HTML]{000000} 0.92773} & 
{\color[HTML]{000000} 565.60429} & 
{\color[HTML]{000000} 1.63656} \\

\midrule

{\color[HTML]{000000} $1$} & 
{\color[HTML]{000000} $1$} & 
{\color[HTML]{000000} $0.6$} & 
{\color[HTML]{000000} $1e-6$} & 
{\color[HTML]{000000} $1$} &
{\color[HTML]{000000} 16.7} &
{\color[HTML]{000000} 0.96148} & 
{\color[HTML]{000000} 0.92773} & 
{\color[HTML]{000000} 565.60429} & 
{\color[HTML]{000000} 1.63656} \\

{\color[HTML]{000000} $1$} & 
{\color[HTML]{000000} $1$} & 
{\color[HTML]{000000} $0.8$} & 
{\color[HTML]{000000} $1e-6$} & 
{\color[HTML]{000000} $1$} &
{\color[HTML]{000000} \textbf{68.3}} & 
{\color[HTML]{000000} \underline{0.17000}} &
{\color[HTML]{000000} \underline{0.05094}} &
{\color[HTML]{000000} \textbf{93.11345}} &
{\color[HTML]{000000} \textbf{0.08429}} \\

{\color[HTML]{000000} $1$} & 
{\color[HTML]{000000} $1$} & 
{\color[HTML]{000000} $1$} & 
{\color[HTML]{000000} $1e-6$} & 
{\color[HTML]{000000} $1$} &
{\color[HTML]{000000} \underline{67.3}} &
{\color[HTML]{000000} \textbf{0.16662}} &
{\color[HTML]{000000} \textbf{0.04666}} &
{\color[HTML]{000000} \underline{98.05333}} &
{\color[HTML]{000000} \underline{0.08452}} \\

{\color[HTML]{000000} $1$} & 
{\color[HTML]{000000} $1$} & 
{\color[HTML]{000000} $2$} & 
{\color[HTML]{000000} $1e-6$} & 
{\color[HTML]{000000} $1$} &
{\color[HTML]{000000} 62.3} & 
{\color[HTML]{000000} 0.20972} & 
{\color[HTML]{000000} 0.07609} & 
{\color[HTML]{000000} 115.81992} & 
{\color[HTML]{000000} 0.10196} \\

{\color[HTML]{000000} $1$} & 
{\color[HTML]{000000} $1$} & 
{\color[HTML]{000000} $3$} & 
{\color[HTML]{000000} $1e-6$} & 
{\color[HTML]{000000} $1$} &
{\color[HTML]{000000} 62.7} & 
{\color[HTML]{000000} 0.18594} & 
{\color[HTML]{000000} 0.06076} & 
{\color[HTML]{000000} 108.61163} & 
{\color[HTML]{000000} 0.09419} \\

\midrule


{\color[HTML]{000000} $1$} & 
{\color[HTML]{000000} $1$} & 
{\color[HTML]{000000} $1$} & 
{\color[HTML]{000000} $1e-8$} & 
{\color[HTML]{000000} $1$} &
{\color[HTML]{000000} 61.7} & 
{\color[HTML]{000000} 0.20459} & 
{\color[HTML]{000000} 0.06836} & 
{\color[HTML]{000000} \underline{109.21708}} & 
{\color[HTML]{000000} 0.09732} \\

{\color[HTML]{000000} $1$} & 
{\color[HTML]{000000} $1$} & 
{\color[HTML]{000000} $1$} & 
{\color[HTML]{000000} $1e-7$} & 
{\color[HTML]{000000} $1$} &
{\color[HTML]{000000} \underline{64.3}} & 
{\color[HTML]{000000} \underline{0.19516}} & 
{\color[HTML]{000000} \underline{0.06514}} & 
{\color[HTML]{000000} 119.81234} & 
{\color[HTML]{000000} \underline{0.09489}} \\

{\color[HTML]{000000} $1$} & 
{\color[HTML]{000000} $1$} & 
{\color[HTML]{000000} $1$} & 
{\color[HTML]{000000} $1e-6$} & 
{\color[HTML]{000000} $1$} &
{\color[HTML]{000000} \textbf{67.3}} &
{\color[HTML]{000000} \textbf{0.16662}} & 
{\color[HTML]{000000} \textbf{0.04666}} & 
{\color[HTML]{000000} \textbf{98.05333}} & 
{\color[HTML]{000000} \textbf{0.08452}} \\

{\color[HTML]{000000} $1$} & 
{\color[HTML]{000000} $1$} & 
{\color[HTML]{000000} $1$} & 
{\color[HTML]{000000} $1e-5$} & 
{\color[HTML]{000000} $1$} &
{\color[HTML]{000000} 50.0} & 
{\color[HTML]{000000} 0.38631} & 
{\color[HTML]{000000} 0.23161} & 
{\color[HTML]{000000} 213.53632} & 
{\color[HTML]{000000} 0.15840} \\

{\color[HTML]{000000} $1$} & 
{\color[HTML]{000000} $1$} & 
{\color[HTML]{000000} $1$} & 
{\color[HTML]{000000} $1e-4$} & 
{\color[HTML]{000000} $1$} &
{\color[HTML]{000000} 16.7} & 
{\color[HTML]{000000} 0.96148} & 
{\color[HTML]{000000} 0.92773} & 
{\color[HTML]{000000} 565.60429} & 
{\color[HTML]{000000} 1.63656} \\

\midrule


{\color[HTML]{000000} $1$} & 
{\color[HTML]{000000} $1$} & 
{\color[HTML]{000000} $1$} & 
{\color[HTML]{000000} $1e-6$} & 
{\color[HTML]{000000} $0.6$} &
{\color[HTML]{000000} 66.0} & 
{\color[HTML]{000000} 0.20317} & 
{\color[HTML]{000000} 0.06447} & 
{\color[HTML]{000000} 106.06283} & 
{\color[HTML]{000000} 0.09527} \\

{\color[HTML]{000000} $1$} & 
{\color[HTML]{000000} $1$} & 
{\color[HTML]{000000} $1$} & 
{\color[HTML]{000000} $1e-6$} & 
{\color[HTML]{000000} $0.8$} &
{\color[HTML]{000000} 16.7} &
{\color[HTML]{000000} 0.96148} & 
{\color[HTML]{000000} 0.92773} & 
{\color[HTML]{000000} 565.60429} & 
{\color[HTML]{000000} 1.63656} \\

{\color[HTML]{000000} $1$} & 
{\color[HTML]{000000} $1$} & 
{\color[HTML]{000000} $1$} & 
{\color[HTML]{000000} $1e-6$} & 
{\color[HTML]{000000} $1$} &
{\color[HTML]{000000} \underline{67.3}} &
{\color[HTML]{000000} \textbf{0.16662}} &
{\color[HTML]{000000} \textbf{0.04666}} & 
{\color[HTML]{000000} \underline{98.05333}} & 
{\color[HTML]{000000} \textbf{0.08452}} \\

{\color[HTML]{000000} $1$} & 
{\color[HTML]{000000} $1$} & 
{\color[HTML]{000000} $1$} & 
{\color[HTML]{000000} $1e-6$} & 
{\color[HTML]{000000} $2$} &
{\color[HTML]{000000} \textbf{69.0}} & 
{\color[HTML]{000000} \underline{0.17601}} & 
{\color[HTML]{000000} \underline{0.05230}} & 
{\color[HTML]{000000} \textbf{97.61372}} & 
{\color[HTML]{000000} \underline{0.08688}} \\

{\color[HTML]{000000} $1$} & 
{\color[HTML]{000000} $1$} & 
{\color[HTML]{000000} $1$} & 
{\color[HTML]{000000} $1e-6$} & 
{\color[HTML]{000000} $3$} &
{\color[HTML]{000000} 16.7} &
{\color[HTML]{000000} 0.96148} & 
{\color[HTML]{000000} 0.92773} & 
{\color[HTML]{000000} 565.60429} & 
{\color[HTML]{000000} 1.63656} \\

\bottomrule
\end{tabular}
\label{pa3}
\end{table*}

\begin{table*}[htbp]
\caption{Parameter analysis on $\hat{T}_{min}$. The best performances are marked \textbf{bold}. The second best performances are marked \underline{underlined}.}
\small
\centering
\begin{tabular}{@{}lccccc@{}}

\toprule

{\color[HTML]{000000} } & 
\multicolumn{1}{c}{{\color[HTML]{000000} Errors $\uparrow$}} &
\multicolumn{4}{c}{{\color[HTML]{000000} Errors $\downarrow$}} \\ 

\cmidrule(l){2-6}
\multirow{-2}{*}{{\color[HTML]{000000} $\hat{T}_{min}$}} &
{\color[HTML]{000000} Accuracy (unit in \%)} &
{\color[HTML]{000000} AbsRel} &
{\color[HTML]{000000} SqRel} &
{\color[HTML]{000000} RMSE} &
{\color[HTML]{000000} RMSElog} \\ 

\midrule

{\color[HTML]{000000} $0$} & 
{\color[HTML]{000000} 21.0} & 
{\color[HTML]{000000} 2.91409} & 
{\color[HTML]{000000} 14.15614} & 
{\color[HTML]{000000} 1140.67145} & 
{\color[HTML]{000000} 0.57624} \\

{\color[HTML]{000000} $1e-3$} & 
{\color[HTML]{000000} 32.3} & 
{\color[HTML]{000000} 0.83141} & 
{\color[HTML]{000000} 1.16434} & 
{\color[HTML]{000000} 433.51288} & 
{\color[HTML]{000000} 0.27273} \\

{\color[HTML]{000000} $1e-2$} & 
{\color[HTML]{000000} \textbf{68.3}} & 
{\color[HTML]{000000} \textbf{0.17000}} &
{\color[HTML]{000000} \underline{0.05094}} &
{\color[HTML]{000000} \textbf{93.11345}} &
{\color[HTML]{000000} \textbf{0.08429}} \\

{\color[HTML]{000000} $1e-1$} & 
{\color[HTML]{000000} \underline{65.7}} & 
{\color[HTML]{000000} \underline{0.17235}} & 
{\color[HTML]{000000} \textbf{0.05026}} &
{\color[HTML]{000000} \underline{102.81039}} & 
{\color[HTML]{000000} \underline{0.08727}} \\

{\color[HTML]{000000} $2e-1$} & 
{\color[HTML]{000000} \underline{65.7}} & 
{\color[HTML]{000000} 0.17609} & 
{\color[HTML]{000000} 0.05288} & 
{\color[HTML]{000000} 103.54889} & 
{\color[HTML]{000000} 0.08893} \\

\bottomrule
\end{tabular}
\label{pa1}
\end{table*}

\begin{table*}[htbp]
\caption{Parameter analysis on $\hat{V}_{max}$. The best performances are marked \textbf{bold}. The second best performances are marked \underline{underlined}.}
\small
\centering
\begin{tabular}{@{}lccccc@{}}

\toprule

{\color[HTML]{000000} } & 
\multicolumn{1}{c}{{\color[HTML]{000000} Errors $\uparrow$}} &
\multicolumn{4}{c}{{\color[HTML]{000000} Errors $\downarrow$}} \\ 

\cmidrule(l){2-6}
\multirow{-2}{*}{{\color[HTML]{000000} $\hat{V}_{max}$}} &
{\color[HTML]{000000} Accuracy (unit in \%)} &
{\color[HTML]{000000} AbsRel} &
{\color[HTML]{000000} SqRel} &
{\color[HTML]{000000} RMSE} &
{\color[HTML]{000000} RMSElog} \\ 

\midrule

{\color[HTML]{000000} $1e3$} & 
{\color[HTML]{000000} 53.3} & 
{\color[HTML]{000000} 0.19826} & 
{\color[HTML]{000000} 0.05809} & 
{\color[HTML]{000000} 147.15601} & 
{\color[HTML]{000000} 0.10736} \\

{\color[HTML]{000000} $1e4$} & 
{\color[HTML]{000000} \underline{68.0}} & 
{\color[HTML]{000000} 0.17071} & 
{\color[HTML]{000000} \textbf{0.05088}} & 
{\color[HTML]{000000} 94.79418} & 
{\color[HTML]{000000} 0.08568} \\

{\color[HTML]{000000} $1e5$} & 
{\color[HTML]{000000} \textbf{68.3}} & 
{\color[HTML]{000000} \textbf{0.17000}} &
{\color[HTML]{000000} \underline{0.05094}} &
{\color[HTML]{000000} \textbf{93.11345}} &
{\color[HTML]{000000} \textbf{0.08429}} \\

{\color[HTML]{000000} $1e6$} & 
{\color[HTML]{000000} \textbf{68.3}} & 
{\color[HTML]{000000} \underline{0.17001}} & 
{\color[HTML]{000000} \underline{0.05094}} & 
{\color[HTML]{000000} \underline{93.11432}} & 
{\color[HTML]{000000} \underline{0.08531}} \\

\bottomrule
\end{tabular}
\label{pa2}
\end{table*}

In this subsection, we conduct some experiments to search for the most suitable set of hyperparameters. 

Table \ref{pa3} shows the results under different $\lambda_A$, $\lambda_T$, $\lambda_D$, $\lambda_{defog}$, and $\lambda_{vis}$. The Control variable method is used to make the searching process achievable. As shown in Table \ref{pa3}, we manually set possible parameters for different hyperparameters and exhaustively conduct experiments to find the suitable set of parameters. Finally, $\lambda_A = \lambda_T = \lambda_{vis} = 1$, $\lambda_D = 0.8$, and $\lambda_{defog} = 1e-6$ are adopted, respectively.

Table \ref{pa1} and \ref{pa2} show the results under different $\hat{T}_{min}$ and $\hat{V}_{max}$. We can learn that setting $\hat{T}_{min}$ and $\hat{V}_{max}$ to $1e-2$ and $1e5$ can achieve the best performance, respectively.


\subsection{Ablation studies}

In this subsection, we conduct some ablation studies on the loss functions $L_{defog}$ and $L_{vis}$; some quantitative results are shown in Table \ref{ablation}. We can learn that solely using $L_{defog}$ will improve the performance, while solely using $L_{vis}$ will lead to the non-convergence of the model. However, simultaneously using $L_{defog}$ and $L_{vis}$ leads to the best performance, which proves the effectiveness and necessity of $L_{defog}$ and $L_{vis}$.

Table \ref{compare2} presents the quantitative comparisons between the proposed model and the naive model; the proposed model integrates the physical laws and the deep learning methods, while the naive model consisting of one Encoder and one Decoder\_T directly estimates the visibility map. The naive model is trained with identical configurations and hyperparameters, and RMSE loss is used during training. Moreover, we take the average value of the visibility map as the image-wise estimate. The numbers presented in Table \ref{compare2} demonstrate that the proposed method outperforms the Naive Model by large margins in all metrics.



\begin{table*}[htbp]
\caption{Ablation studies on loss function. The best performances are marked \textbf{bold}. The second best performances are \underline{underlined}.}
\small
\centering
\begin{tabular}{@{}ccccccc@{}}

\toprule

{\color[HTML]{000000} } & 
{\color[HTML]{000000} } & 
\multicolumn{1}{c}{{\color[HTML]{000000} Errors $\uparrow$}} &
\multicolumn{4}{c}{{\color[HTML]{000000} Errors $\downarrow$}} \\ 

\cmidrule(l){3-7}
\multirow{-2}{*}{{\color[HTML]{000000} $L_{defog}$}} &
\multirow{-2}{*}{{\color[HTML]{000000} $L_{vis}$}} &
{\color[HTML]{000000} Accuracy (unit in \%)} &
{\color[HTML]{000000} AbsRel} &
{\color[HTML]{000000} SqRel} &
{\color[HTML]{000000} RMSE} &
{\color[HTML]{000000} RMSElog} \\ 

\midrule

{\color[HTML]{000000} -} & 
{\color[HTML]{000000} -} & 
{\color[HTML]{000000} 60.0} & 
{\color[HTML]{000000} 0.21444} & 
{\color[HTML]{000000} 0.08606} & 
{\color[HTML]{000000} 109.64792} & 
{\color[HTML]{000000} 0.10298} \\

{\color[HTML]{000000} -} & 
\multicolumn{1}{c}{{\color[HTML]{000000} $\surd$}} &
{\color[HTML]{000000} 16.7} &
{\color[HTML]{000000} 0.96148} & 
{\color[HTML]{000000} 0.92773} & 
{\color[HTML]{000000} 565.60429} & 
{\color[HTML]{000000} 1.63656} \\

\multicolumn{1}{c}{{\color[HTML]{000000} $\surd$}} &
{\color[HTML]{000000} -} & 
{\color[HTML]{000000} \underline{67.3}} & 
{\color[HTML]{000000} \underline{0.21279}} & 
{\color[HTML]{000000} \underline{0.08279}} & 
{\color[HTML]{000000} \underline{107.93496}} & 
{\color[HTML]{000000} \underline{0.10204}} \\

\multicolumn{1}{c}{{\color[HTML]{000000} $\surd$}} &
\multicolumn{1}{c}{{\color[HTML]{000000} $\surd$}} &
{\color[HTML]{000000} \textbf{68.3}} & 
{\color[HTML]{000000} \textbf{0.17000}} &
{\color[HTML]{000000} \textbf{0.05094}} &
{\color[HTML]{000000} \textbf{93.11345}} &
{\color[HTML]{000000} \textbf{0.08429}} \\

\bottomrule
\end{tabular}
\label{ablation}
\end{table*}

\begin{table*}[htbp]
\caption{Quantitative results of the visibility estimation on the FACI dataset. The best performances and our model are marked \textbf{bold}.}
\small
\centering
\begin{tabular}{@{}lccccc@{}}

\toprule

{\color[HTML]{000000} } & 
\multicolumn{1}{c}{{\color[HTML]{000000} Errors $\uparrow$}} &
\multicolumn{4}{c}{{\color[HTML]{000000} Errors $\downarrow$}} \\ 

\cmidrule(l){2-6}
\multirow{-2}{*}{{\color[HTML]{000000} Methods}} &
{\color[HTML]{000000} Accuracy (unit in \%)} &
{\color[HTML]{000000} AbsRel} & 
{\color[HTML]{000000} SqRel} & 
{\color[HTML]{000000} RMSE} & 
{\color[HTML]{000000} RMSElog} \\ 

\midrule

{\color[HTML]{000000} Naive Model} & 
{\color[HTML]{000000} 55.3} &
{\color[HTML]{000000} 0.21116} & 
{\color[HTML]{000000} 0.06993} & 
{\color[HTML]{000000} 143.15237} & 
{\color[HTML]{000000} 0.11181} \\

{\color[HTML]{000000} \textbf{Our} } &
{\color[HTML]{000000} \textbf{68.3}} & 
{\color[HTML]{000000} \textbf{0.17000}} &
{\color[HTML]{000000} \textbf{0.05094}} &
{\color[HTML]{000000} \textbf{93.11345}} &
{\color[HTML]{000000} \textbf{0.08429}} \\

\bottomrule
\end{tabular}
\label{compare2}
\end{table*}

\subsection{Model complexity}

Table \ref{complexity} summarizes the model’s time and space complexities. We can learn that the time and space complexities of the proposed method are on a par with other state-of-the-art methods. However, the performance of the proposed method outperforms those of state-of-the-art methods by large margins as presented in Table \ref{compare}.


\begin{table}[ht]
\caption{Model complexity. The time complexity evaluations are performed at an image resolution of 288 × 512 on a single GeForce GTX 1080. TC and SC represent time and space complexities, respectively. The best performances and our model are marked \textbf{bold}. The second best performances are \underline{underlined}.}
\small
\centering
\begin{tabular}{lcc}
\toprule
Methods & TC (GFLOPs) & SC (M) \\  \midrule
VGG-16 \cite{simonyan2014very} & 42.23 & 134.28 \\
ResNet-50 \cite{he2016deep} &  \textbf{12.18} & 23.52 \\ 
VisNet \cite{palvanov2019visnet} &  37.92 & \textbf{16.04} \\ 
\textbf{Our} & \underline{14.10} & \underline{19.30} \\
\bottomrule
\end{tabular}
\label{complexity}
\end{table}

\section{Limitations}

In this section, the limitations of the proposed model are discussed. Firstly, the method requires the depth map and the transmission map as supervision to train the proposed framework, which may limit the use of the method. Secondly, for the images excluding sky area, the airlight may not be well estimated, further degenerating the performance of the proposed method. We will work on these issues in the future.



\section{Conclusion}

Visibility estimation under foggy weather is important to traffic safety and transportation infrastructure systems management. However, most methods mainly adopt professional instruments outfitted at fixed positions on the road to estimate visibility, which is costly and less flexible. Although many trials applying deep learning methods to visibility estimation have been conducted, the estimation accuracy still needs to be improved. Importantly, none of the previous methods considers the uneven fog situations, which occurs frequently in really, especially in mountain areas. 

To mitigate these research gaps, we propose an innovative visibility estimation framework, which integrates deep neural networks and physical law to improve performance. Moreover, we, for the first time, propose a pixel-wise visibility estimation paradigm to take the uneven fog situations into account, which is more informative and practical than a single-valued estimation. Finally, a new virtual dataset, FACI, is collected and proposed for validating the proposed method. Detailed experiments on the FACI dataset and the real-world data demonstrate the effectiveness and practicality of the proposed method. We hope that the proposed method can contribute to the development of intelligent transportation infrastructure systems.

\ifCLASSOPTIONcaptionsoff
  \newpage
\fi



%

\bibliography{IEEEabrv,IEEEtran.bib}{}
\bibliographystyle{IEEEtran}

\end{document}